\documentclass[letterpaper, 10 pt, conference]{ieeeconf}  
\IEEEoverridecommandlockouts                            
\usepackage{graphics} % for pdf, bitmapped graphics files
\usepackage{epsfig} % for postscript graphics files
\usepackage{mathptmx} % assumes new font selection scheme installed
\usepackage{times} % assumes new font selection scheme installed
\usepackage{amsmath} % assumes amsmath package installed
\usepackage{amssymb}  % assumes amsmath package installed
\usepackage{url}
\begin{document}
\title{\LARGE \bf
Underactuated multimodal jumping robot for extraterrestrial exploration
}

\author{Neil R. Wagner$^{1}$ and Justin K. Yim$^{1}$% 
\thanks{The material is based upon work supported by NASA under award No 80NSSC25K7609}% 
\thanks{$^{1}$Department of Mechanical Science and Engineering,
University of Illinois Urbana-Champaign
        {\tt\small neilrw2@illinois.edu} {\tt\small jkyim@illinois.edu}}%
}
\maketitle
\thispagestyle{empty}
\pagestyle{empty}
\begin{abstract}
We present a rolling and jumping underactuated monopedal robot designed to explore multimodal locomotion on low-gravity bodies. It uses only two reaction wheels to control its spatial orientation with two controllers: a balancing controller which can aim the robot's jump direction on the ground, and an aerial reorientation controller which can aim the robot's leg for landing after flight. We demonstrate rolling, targeted jumping and landing, and self-righting using only three actuators total, keeping system size to $0.33~m$ and $1.25~kg$. Simple switching between locomotion modes enables the system to deal with differing landscapes and environmental conditions.
\end{abstract}

\section{INTRODUCTION}
In the search for life on other bodies in the solar system, Saturn's moon Enceladus is a key candidate. Its subsurface liquid water ocean with organic molecules points to a potentially habitable environment for life like that on Earth. To study this environment, a robotic mission could collect samples from the cryovolcanic jets coming through large fissures near the south pole. However, exploring this frozen moon presents a variety of challenges for conventional space robots. Fine ice particles and hundred meter high ridges make up the surface near the jets.  The body exhibits a gravity $1/80^{th}$ as strong as Earth's and has no atmosphere. Additionally, the average temperature on the surface is around $-200^\circ~C$.

Despite these challenges, exploring this exotic environment remains a key focus for space associations, due to the immense potential for scientific discovery \cite{cable_science_2021,mackenzie_enceladus_2021,des_marais_nasa_2008,choukroun_sampling_2021}. 
Proposed missions like Enceladus Orbilander plan to sample plume material in orbit and from a fixed location on the surface, but lack a mechanism for sampling material directly at the jets \cite{mackenzie_enceladus_2021}. Such measurements taken by a small robot accompanying a larger lander could investigate plume eruption mechanics and makeup. Traveling greater distances to sample multiple jets could also identify variation among jets \cite{vaquero2024eels}.

In this paper, we propose an underactuated monopedal jumping and rolling robot that aims to address these challenges. On flat surfaces, it can roll like a typical differential drive mobile robot using its wheels. On granular media and uneven terrain, the robot can stand up and balance on its single leg, as well as lean and jump in desired directions. In flight, the robot can reorient itself with its reaction wheels and point its leg in any direction to land precisely. This robot could be uniquely suited to explore small bodies like Enceladus where the low gravity enables enormous jumps to cross gaps and fissures, collecting data and surveying the landscape. Using the two locomotion modes, it could provide an efficient and versatile method of traveling on this distant moon.

Relatively few terrestrial robots possess the ability to balance on a point in 3D and reorient in flight. For a monopedal jumping robot, these are both necessary to jump reliably and efficiently. In this paper, we introduce methods of balancing and aerial reorientation with only two reaction wheels. Using two reaction wheels reduces system complexity and weight to improve robustness and decrease power consumption.  This innovation is particularly useful where power usage must be low and in non-atmospheric environments where aerodynamic forces are useless. A prime example of this environment is space missions on other bodies where atmospheres are absent and energy reserves are limited. 

\begin{figure}
    \centering
    \includegraphics[width=.75\linewidth]{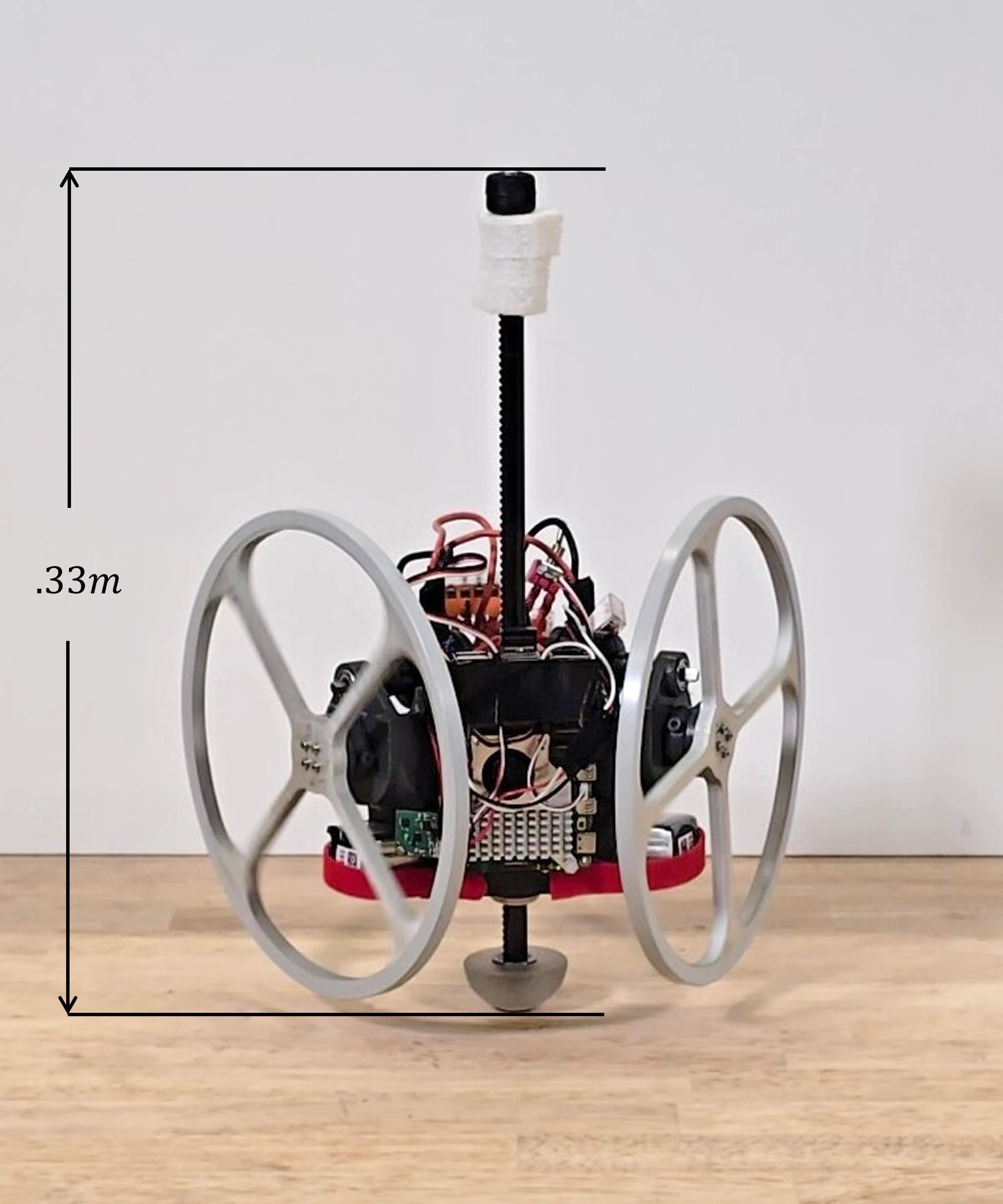}
    \caption{Underactuated robot with one leg and two reaction wheels balancing on a table}
    \label{fig:placeholder}
\end{figure}
    
Prior research has often used reaction wheels for balancing, like in CUBLI \cite{gajamohan_cubli_2012,muehlebach_nonlinear_2017}, and aerial reorientation, such as in satellites \cite{rui_nonlinear_2000} and hopping robots \cite{nomura_attitude_2018}, but has rarely approached the problem with fewer than three actuators to control all directions. In addition, while some robots have used different actuation methods like propellers, such as Hopcopter \cite{burns_design_2025} or Salto \cite{haldane_repetitive_2017}, or extra appendages \cite{chu_combining_2023,zelinka_attitude_2020,libby_comparative_2016,jianguo_zhao_controlling_2013}, they still use actuation over all three dimensions. Additionally, rocket or jet-based systems are unsuitable for this exploration as exhausted propellant could contaminate samples and data gathered. 

Precise control of jumping and landing allows this robot to reach areas that would otherwise be unattainable by other locomotion methods. While other robot platforms have demonstrated jumping behaviors such as reaction wheel launching, like Hedgehog \cite{kulic_experimental_2017}, and leg-based jumping \cite{kalita_dynamics_2020}, they lack the ability to control their landing, which leaves them susceptible to becoming easily stuck or lost on steep slopes and crevasses.
    
Due to unknown landscapes and changing environmental conditions, the ability to operate reliably is required. Other planetary and lunar exploration robots may be limited while encountering unknown and changing conditions, such as the Mars rovers getting stuck in terrain \cite{david2005opportunity}. Previous research has explored jumping and rolling robots such as Boston Dynamics' Sand Flea \cite{ackerman2012boston} which cannot control its landing, or  ETH Zurich's Ascento \cite{klemm_ascento_2019}, which needs a large solid surface to operate. There have even been concepts for jumping robots in space, like ETH Zurich's Space Hopper \cite{spiridonov_spacehopper_2024}, however they require more actuation, increasing complexity, and lack the ability to roll, reducing capability. 
    
This paper expands planar balancing control presented in \cite{featherstone_quantitative_2016,featherstone_simple_2017}. By applying this controller in two separate planes, spatial balance can be achieved. Since our robot has only two orientation actuators, it cannot fully actuate all orientation directions -- specifically, it cannot directly control yaw about the vertical axis.  However, the robot can accommodate non-zero angular velocities in this direction both in balancing control and aerial reorientation control. Since the jumping propulsion is not dependent on direction and rolling can occur on both sides of the robot, control over yaw while balancing is not required for this system.
    
This paper also introduces an aerial controller. It drives the robot to point is leg along a desired vector, even while dealing with arbitrary non-zero angular momentum. This allows the system to handle varying launch conditions and disturbances.

Lastly, the robot introduced can use both of these controllers to balance on a point, launch itself in the air, reorient itself to aim the leg in the direction to prepare for landing or for consecutive jumps. If the system tips over intentionally or unintentionally, it can right itself and begin balancing.

\begin{figure}
    \centering
    \includegraphics[width=0.75\linewidth]{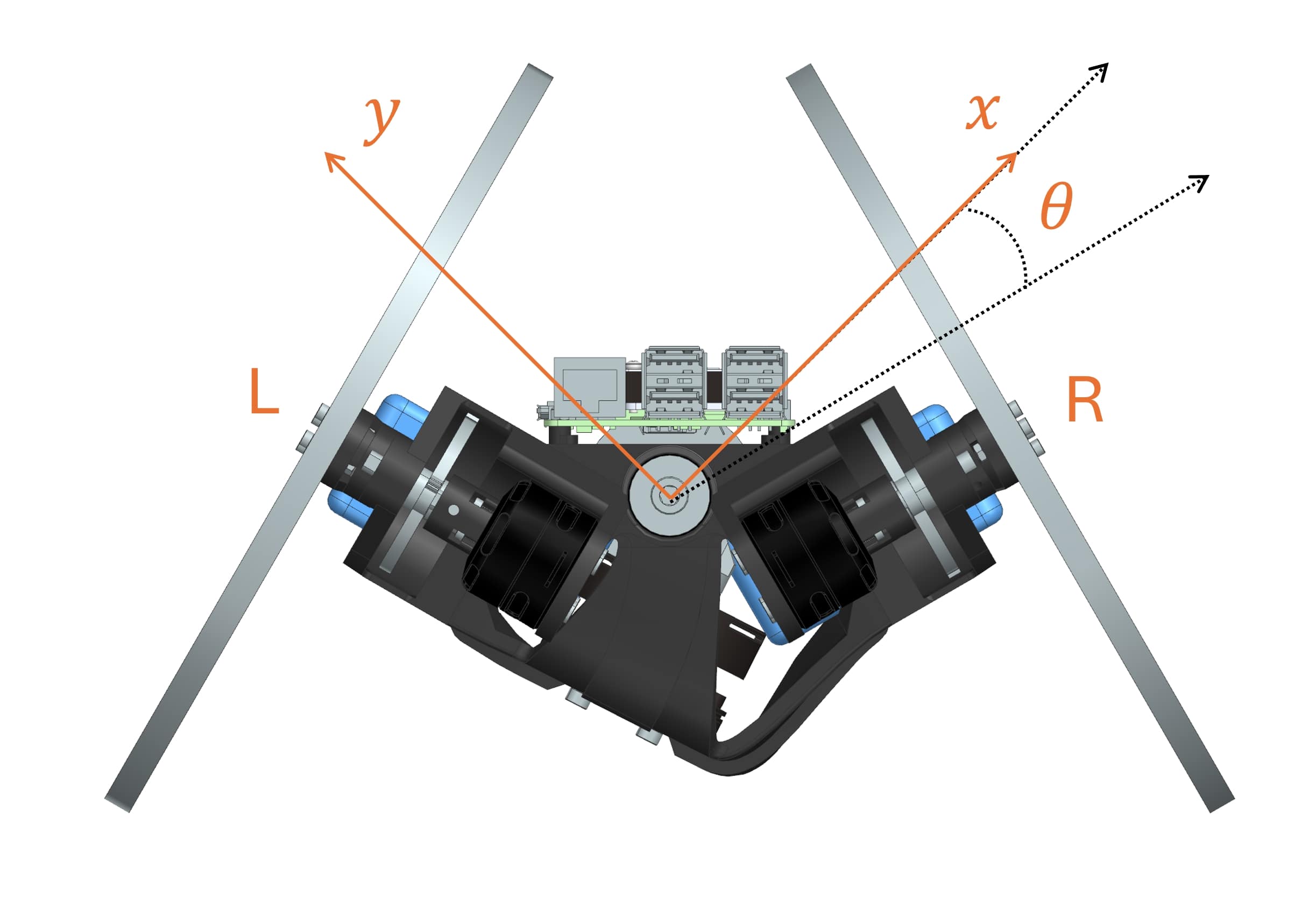}
    \caption{Robot platform from above. The 15-degree cant angle, $\theta$, of the wheels allows for balancing and aerial reorientation while maintaining differential drive features.}
    \label{1}
\end{figure}\begin{figure}
    \centering
    \includegraphics[width=0.75\linewidth]{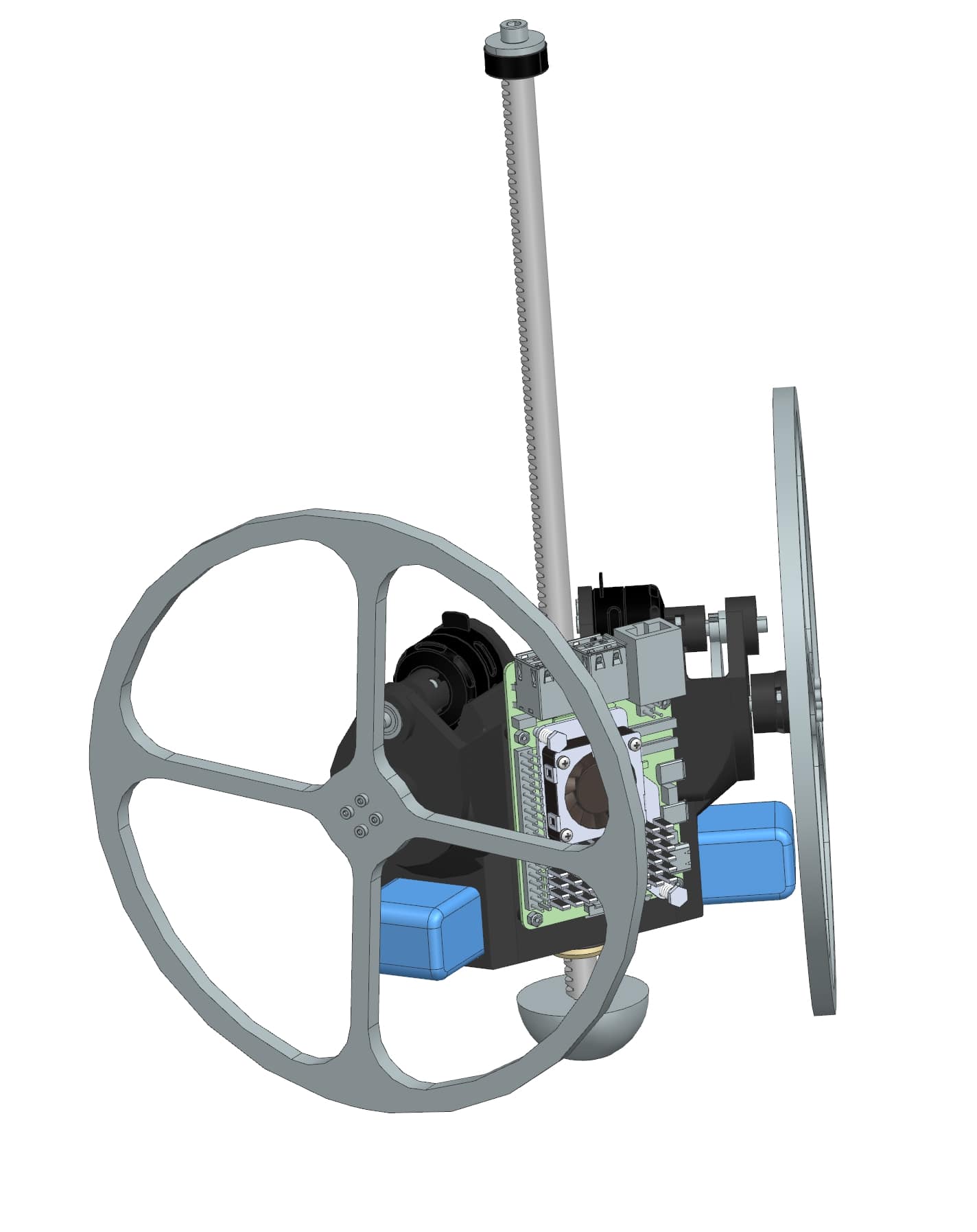}
    \caption{Isometric view of robot platform displayed in CAD}
    \label{2}
\end{figure}
\begin{figure}
    \centering
    \includegraphics[width=0.25\linewidth]{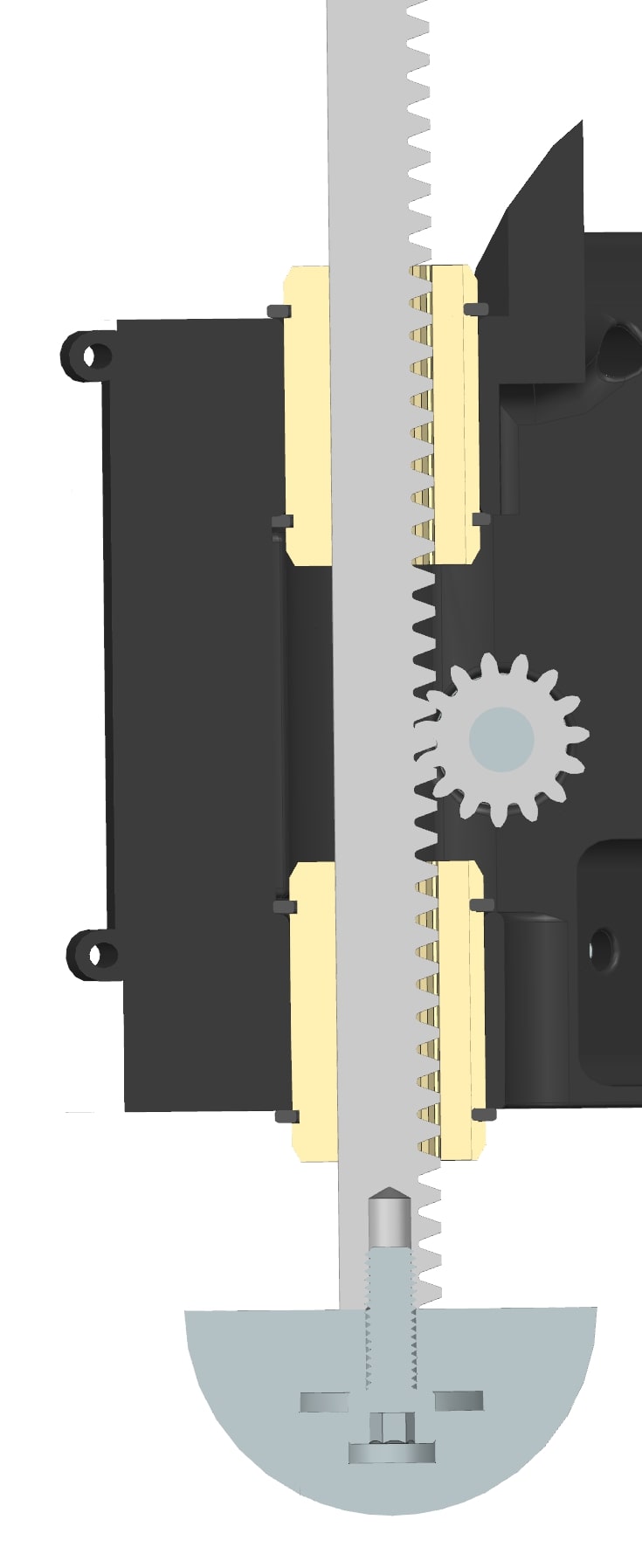}
    \caption{Rack and pinion design for jumping locomotion}
    \label{3}
\end{figure}

\section {METHODS}
We develop control for a robot consisting of one jumping leg and two wheels. The leg extends the foot in a straight line along the z-axis of the robot's chassis-fixed frame approximately coincident with the robot's center of mass. The foot is small and is modeled as a point contact with the ground. The two wheels' axes are orthogonal to the line of action of the leg and are angled relative to each other by a cant angle shown in Fig \ref{1}. When balancing and in flight, these act as reaction wheels to apply torques to the body and when rolling on the ground they operate as usual terrestrial wheels. 
        \subsection{Aerial controller}
In flight the robot must reorient from its launch attitude so that it can land on its foot. To do this, the robot should be able to point its foot in any direction while accommodating residual angular momentum retained at takeoff. This is particularly difficult with only two reaction wheels since they leave the robot underactuated. Previous robots achieve in-flight foot pointing in varying ways, such as using extra actuators as in Salto-1P directly actuating the yaw direction to counteract any non-zero angular momenta encountered at launch \cite{haldane_repetitive_2017}. Other jumping robots use a two degree of freedom servo joint at the base of the leg, like the Raibert Hopper \cite{raibert_hopping_1984} or the 3D bow-legged hopper \cite{zeglin_first_nodate}. This method would severely limit the range of motion that our leg could aim in and require additional actuators, increasing complexity and power consumption. Using the reaction wheels that will also be used in the balancing and rolling modes, we can increase the capability of our underactuated system without adding additional components. 

The aerial controller commands motor torques for the left and right reaction wheel $\tau_R$ and $\tau_L$ respectively in order to drive the vector from the robot CG to the foot (along the leg axis) to align with a desired unit vector $v^0_{foot}$ with respect to the world-fixed frame 0 .
Since we are interested in controlling the direction of the leg, rotation around the axis of the leg during flight is acceptable since it does not change the leg's direction.
Rotation about the leg axis may be induced by nonzero angular momentum at liftoff remaining from the dynamics or disturbances during launch.
Thus, we first derive reaction wheel torque commands that cancel out gyroscopic precession induced by angular momentum of the reaction wheels in order to stabilize the leg pointing axis.  Since the wheels are designed to have identical inertia tensors and spin about the tensor's major axis, the compensatory torque commands are given by:

$$    
\begin{matrix}
\tau_{gyro,R} = I_{wheel}\cdot\omega_{R}\cdot\omega_{b,z}^b\cdot\tan(2\theta)\\
+  I_{wheel}\cdot\omega_{L}\cdot\omega_{b,z}^b/\cos( 2\theta)
\end{matrix}\eqno{(1a)}
$$
$$    
\begin{matrix}
\tau_{gyro,L} = I_{wheel}\cdot\omega_{L}\cdot\omega_{b,z}^b\cdot\tan(2\theta)\\
-  I_{wheel}\cdot\omega_{R}\cdot\omega_{b,z}^b/\cos( 2\theta)
\end{matrix}\eqno{(1b)}
$$
where $\omega_{b,z}^b$ is the component of the angular velocity of the chassis relative to the world frame about the body-fixed z-axis, $I_{wheel}$ is the moment of inertia of a wheel about its axle, $\omega_{R}$ and $\omega_L$ are the angular velocities of the left and right reaction wheels about their axles respectively, and $\theta$ is the cant angle as shown in Fig \ref{1}.

After applying the precession cancellation torques, we can linearize the rotational dynamics about the desired pointing vector to derive a simple proportional-derivative feedback law.  Expressing the desired pointing vector with respect to the body frame $v_{foot}^b = R_0^b v_{foot}^0$ using rotation matrix $R_0^b$ from the world frame to the body frame yields the following expression for the controller:

$$
\begin{matrix}
\tau_{R} = -k_p\cdot v_{foot,y}^b + k_d\cdot\omega_{b,x}^b + \tau_{gyro,R} \\
\tau_{L} = k_p\cdot v_{foot,x}^b + k_d\cdot\omega_{b,y}^b + \tau_{gyro,L}
\end{matrix}
\eqno{(2)}
$$
where $\cdot^b_{\cdot,x}$ and $\cdot^b_{\cdot,y}$ denote components of vectors relative to the body-fixed x and y axes respectively.

        \subsection{Balance controller}
To address balancing on the ground, we propose using two planar controllers, given in previous experiments \cite{driessen_experimental_2019}, in the roll and pitch directions, respectively. To do this with two rotating masses that are not aligned with the coordinated planes aligned with roll and pitch, we combine the effects of each rotating mass such that they are coordinated for each controller in their component directions. The equations of motion of a planar reaction wheel pendulum given by the controller \cite{driessen_experimental_2019}, is
$$
\begin{bmatrix}
H_{11} & H_{12}\\
H_{21} & H_{22}
\end{bmatrix} 
\begin{bmatrix}
\ddot q_ {b}\\
\ddot q_{w} 
\end{bmatrix} 
=
\begin{bmatrix}
mcg\sin(q_{b})\\
\tau 
\end{bmatrix}\eqno{(3)}
$$
where $H_{ij}$ are the elements of the joint space inertia matrix, $\ddot q$, are the joint acceleration variables with subscript $w$ indicating the reaction wheel, and $b$ indicating the body. $\tau$ is the joint torque that acts on the reaction wheel. $c$ is the distance from the foot to the center of mass of the robot, $m$ is the mass of the robot, and $g$ is the acceleration due to gravity. For the 3D robot proposed, the equations of motion that we use for balance controller derivation are
$$
\setlength{\arraycolsep}{1.2pt} 
\begin{bmatrix}
H_{11} & 0&H_{13} & H_{14}\\
0 & H_{22}&H_{23} & H_{24}\\
H_{31} & H_{32}&H_{33} & H_{34}\\
H_{41} & H_{42}&H_{43} & H_{44}
\end{bmatrix} 
\begin{bmatrix}
\ddot q_ {x}\\
\ddot q_{y}\\
\ddot q_ {R}\\
\ddot q_{L} 
\end{bmatrix} 
= 
\begin{bmatrix}
mcg\sin(q_{x})\cos(q_{y})\\
mcg\sin(q_{y})\cos(q_{x})\\
\tau_{R,w} \\
\tau_{L,w}
\end{bmatrix}
\eqno{(4)}$$

where $q_x$ and $q_y$ refer to the angle of the chassis's tilt from vertical about the corresponding axis. Furthermore, as in the original controller, $\ddot q_{x}$ and $\ddot q_{y}$ can be replaced by the non-dimensionalized triple derivative of the angular momentum of the body about the foot's point of contact with the ground, $\dddot M_{x}$ and $\dddot M_{y}$. Rearranging the matrices, one can solve for the joint accelerations for the reaction wheels
$$
\setlength{\arraycolsep}{1.2pt} 
\begin{bmatrix}
H_{13} & H_{14}\\
H_{23} & H_{24}
\end{bmatrix} ^{-1}
\begin{bmatrix}
mcg\sin(q_{x})\cos(q_{y})-\dddot M_xH_{11}\\
mcg\sin(q_{y})\cos(q_{x})-\dddot M_yH_{22}
\end{bmatrix} 
=
\begin{bmatrix}
\ddot q_ {R}\\
\ddot q_ {L}
\end{bmatrix} \eqno{(5)}$$
and therefore solve for the reaction wheel joint torques obtaining
$$
\setlength{\arraycolsep}{1.2pt} 
\begin{bmatrix}
H_{33} & H_{34}\\
H_{43} & H_{44}
\end{bmatrix} 
\begin{bmatrix}
\ddot q_ {R}\\
\ddot q_ {L}
\end{bmatrix} 
+
\begin{bmatrix}
\dddot M_xH_{31}+\dddot M_yH_{32}\\
\dddot M_xH_{41}+\dddot M_yH_{42}
\end{bmatrix} 
=
\begin{bmatrix}
\tau_ {R}\\
\tau_ {L}
\end{bmatrix} \eqno{(6)}$$
Now the two planar balance controllers use two sets of commands given by the control law to balance in 3D space on a point.
$$
\begin{matrix}
\dddot M_{x} = k_{dd}\ddot M_{x}+k_d\dot M_{x} + k_MM_{x}\\
\dddot M_{y} = k_{dd}\ddot M_{y}+k_d\dot M_{y} + k_MM_{y}
\end{matrix} \eqno{(7)}$$
The gains are found using the pole placement method discussed in the planar case \cite{featherstone_simple_2017}, eq. 31-33. Using the small-angle approximation, $M_{x,y}$, $\dot M_{x,y}$, and $\ddot M_{x,y}$ become $T_{c,x,y}^2(\dot q _{x,y}-G_{\omega,x,y} \dot q_{R,L} -G_{\omega,y,x} \dot q_{L,R})$,     $q_{x,y}$,  and $\dot q_{x,y}$, respectively, with $T_{c,x,y}$ being the time constant of toppling and $G_{\omega,x,y}$ being the velocity gain in their respective planes.

        \subsection{Hardware Platform}
The reaction wheels are driven by 4:1 gearboxes powered by 23-06 220Kv Vertiq motors. The reaction wheels also serve as the differential drive wheels while the robot is in the rolling mode on its side. 

The leg is a rack and pinion driven by a 40-06 370Kv Vertiq motor, depicted in Fig. \ref{3}. We use a 300mm long circular steel rack and 1.5 cm diameter 15 tooth pinion. The simple rack and pinion, instead of a linkage-based design of other robotic jumping platforms \cite{kovac_miniature_2008,zhao_miniature_2014,haldane_repetitive_2017}, allows evaluation of custom torque profiles for the jump mechanism. In future work, we plan to simulate candidate linkage mechanisms to evaluate which are most suitable. The rack and pinion leg stores no elastic energy on landing, making our design less efficient but more versatile for evaluating the future design of the leg.

An onboard Raspberry Pi 5, 8GB RAM, handles  all computations. An unscented Kalman filter fuses accelerometer and gyroscope readings from a 6-axis MPU6050 Inertial Measurement Unit (IMU) to estimate the robot's attitude. Since the center of mass is not directly over the leg of the robot, a balance offset observer adjusts the balance point of the controller, as introduced in \cite{driessen_experimental_2019}, eq. 24. Two 3S Li-ion batteries in series with a nominal voltage 22V power all three motors and the computer.

The robot's program runs through a state machine that directs which controller to use on both the leg and reaction wheels. The onboard sensing and control loop fusing IMU and encoder data and commanding motor torques runs at 500 Hz.

\section{RESULTS}
To validate the abilities of the robotic platform, we ran a series of experimental hardware tests which combined maneuvers and controllers to explore the capabilities and effectiveness of the methods presented.
 
 \begin{figure}
    \centering
    \includegraphics[width=1\linewidth]{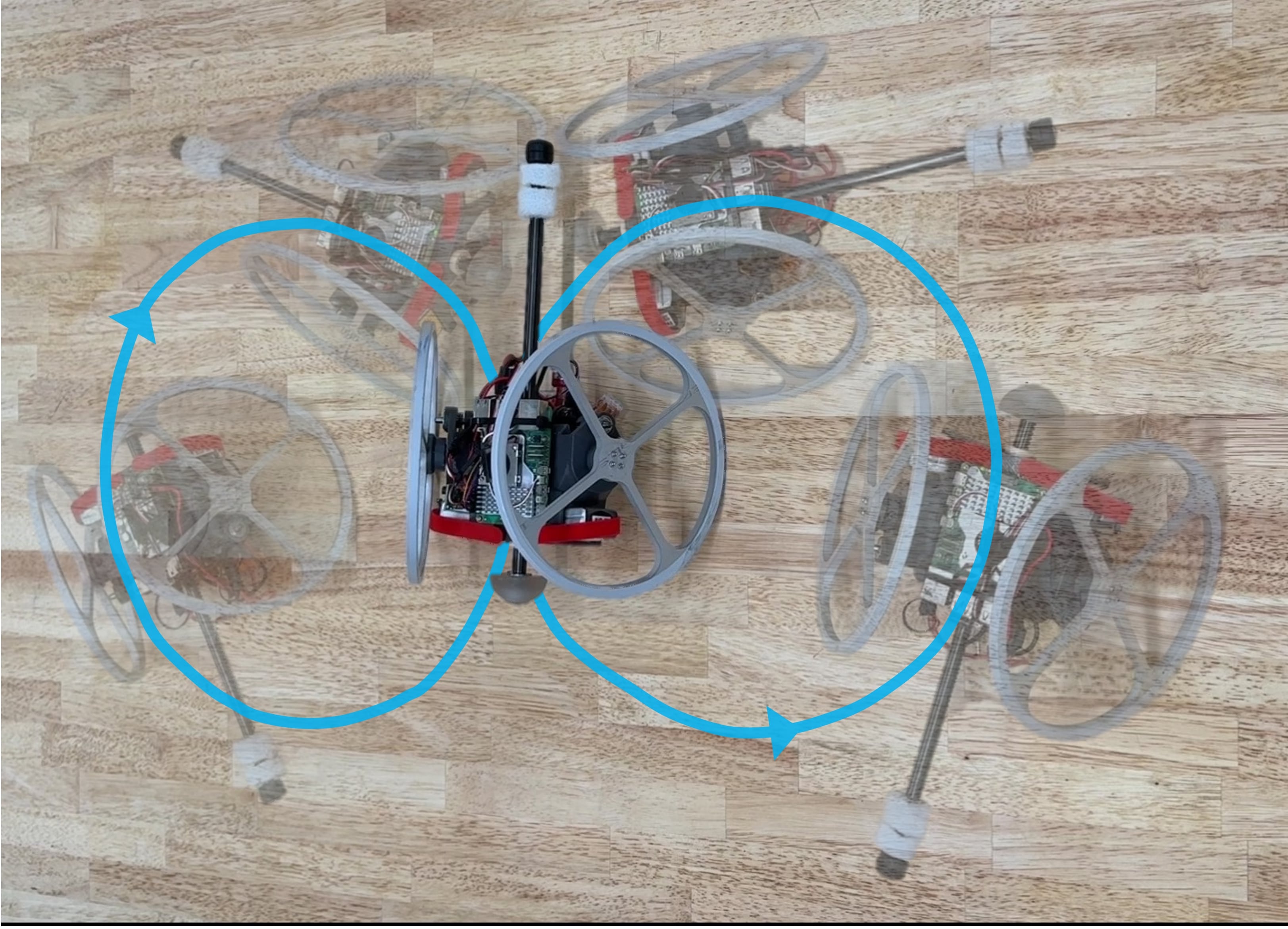}
    \caption{Robot rolls on the ground in a figure-eight path, blue trajectory overlaid from camera tracking.}
    \label{4}
\end{figure}
\begin{figure}
    \centering
    \includegraphics[width=1\linewidth]{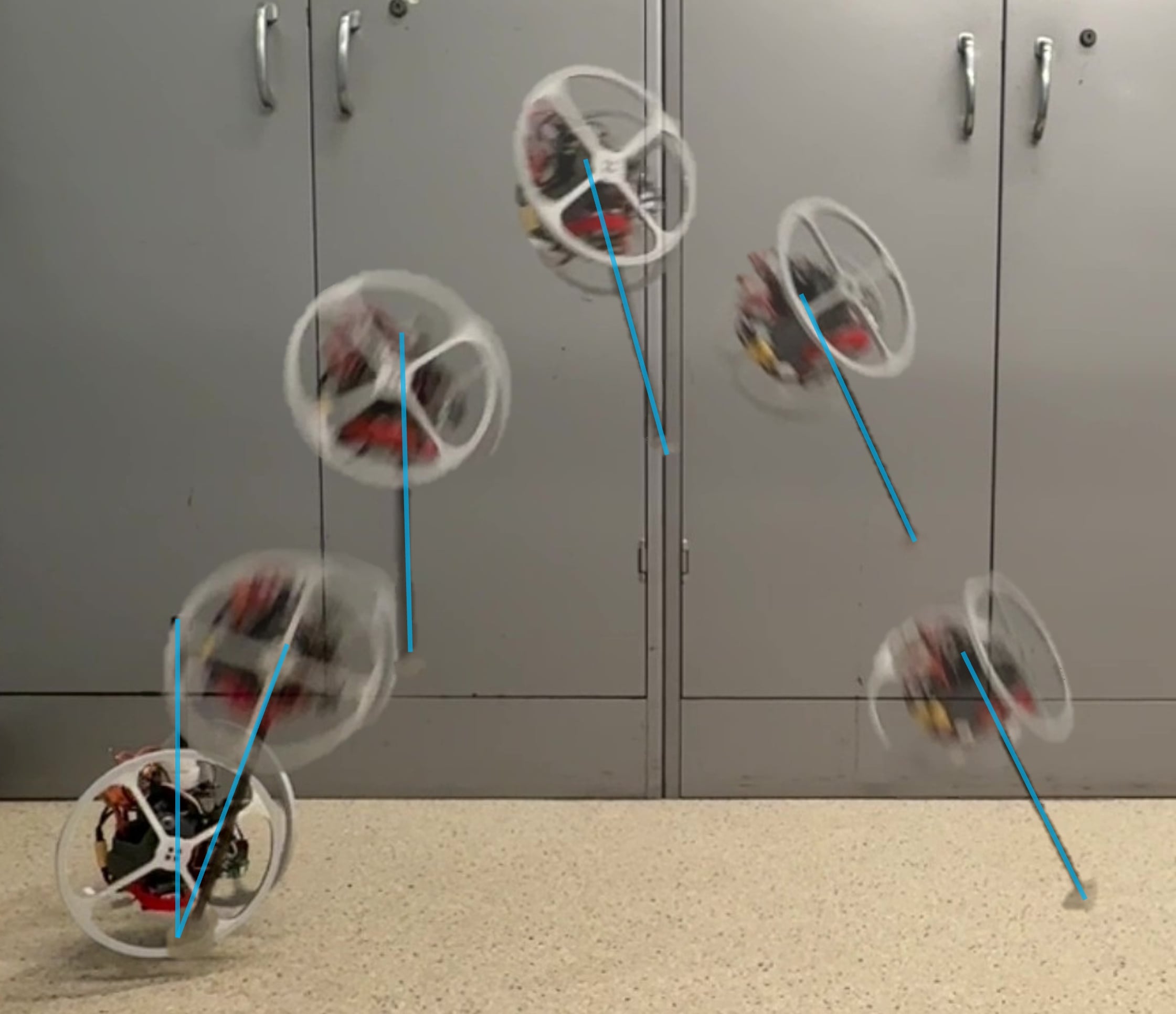}
    \caption{Robot leaning and jumping, then reorientating itself in the air to point along the velocity vector. The data collected in this test is shown in Fig. \ref{6} and Fig. \ref{7}.}
    \label{5}
\end{figure}

\subsection{Rolling in a figure eight}
On its side, the robot can roll and steer like a differential drive robot. Shown in figure \ref{4}, the robot rolls in a figure eight shape showing ability to control its path on the ground. The robot is able to drive when flipped over on the side with the narrower wheel base but usually drives on the side with the wider wheel base for greater stability. The robot can follow a pre-planned trajectory as shown here or accept input from a human user through a joystick.

\begin{figure}
    \centering
    \includegraphics[width=1\linewidth]{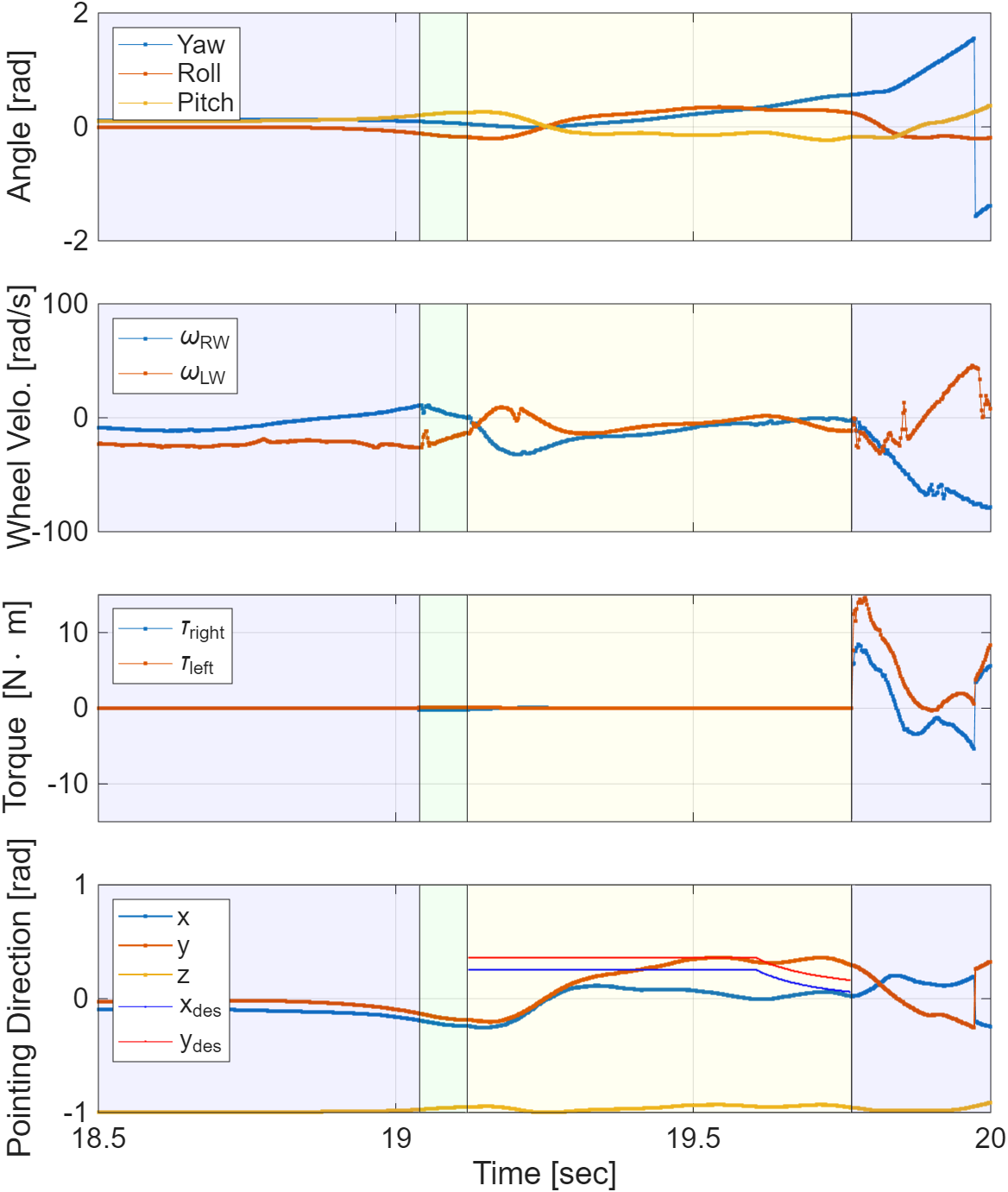}
    \caption{Data traces from the lean jump depicted in figure \ref{5}. Chassis orientation Euler angles, wheel velocity and torques, and leg pointing direction are displayed. The colored regions indicate the change in robot state: blue is balancing, green is jumping, and yellow is aerial.}
    \label{6}
\end{figure}
\begin{figure}
    \centering
    \includegraphics[width=1\linewidth]{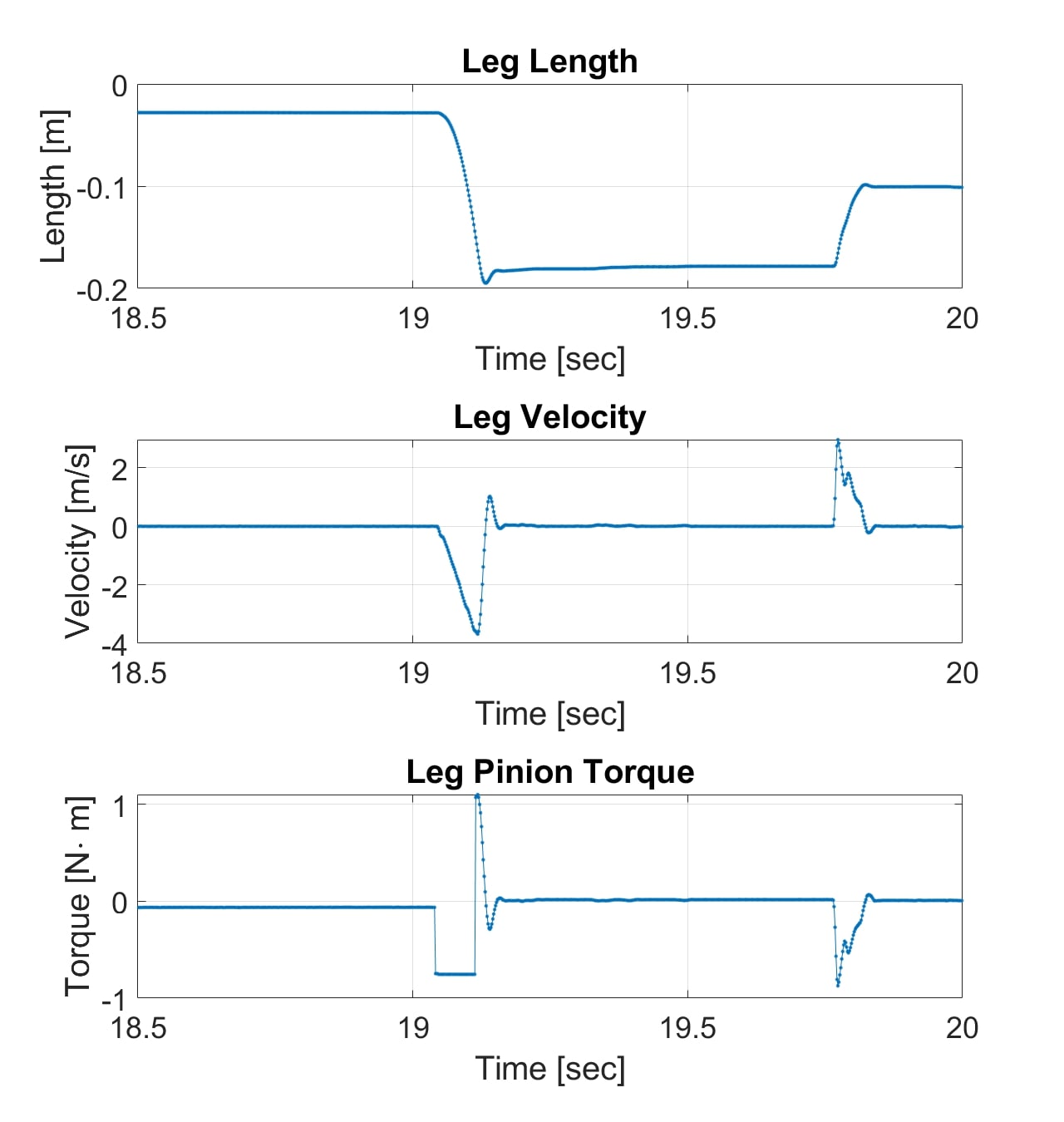}
    \caption{Leg position, velocity, and commanded torque from the leg motor. The first peak is the robot jumping by extending the rack and pinion. The second peak is the landing as the leg motor provides a counter torque to dampen the impact force.}
    \label{7}
\end{figure}

\subsection{Horizontal jump}
While balancing, the balance controller can lean the robot to ``point" towards a desired jump target. Upon reaching the desired angle, the leg motor quickly accelerates the robot to make a lateral jump. During launch, leg encoder and IMU angular velocity measurements are fused to estimate the achieved liftoff velocity vector.  After liftoff, the aerial controller aims the robot's leg along the robot's velocity vector in preparation for landing, though during the initial portion of flight, the leg angle is limited not to point above a threshold angle since landing is expected only after apex when the robot's vertical velocity is negative. This aligns the leg so that after touchdown it can dissipate the energy of the jump using an impedance controller simulating a damper. The landing dissipation allows for a soft tumble, from which the robot can continue rolling or balancing immediately.
Shown in Fig. \ref{6}, the robot launches at an angle of $20^\circ$ from upright, jumping $0.59~m$ high and $0.82~m$ long and the aerial controller aligns the robot along the velocity vector as the robot falls. After touchdown, the robot dampens the impact from the jump to land softly as shown in Fig. \ref{7}. Note that on Enceladus with its weaker gravity, this jump performance would enable jumps over $40~m$ high and $60~m$ long.
Since the launch and landing acceleration magnitudes are much greater than gravity on Earth, gravity can be considered a minor disturbance in the dynamics during the high-acceleration phases of motion and their behavior in lower gravity environments would be similar.  However, residual velocity that results in a bounce after landing could cause longer flight times in lower gravity; in this case, aerial control could be again re-engaged for subsequent attempts at landing or the robot could self-right after coming to rest.

\begin{figure}
    \centering
    \includegraphics[width=0.75\linewidth]{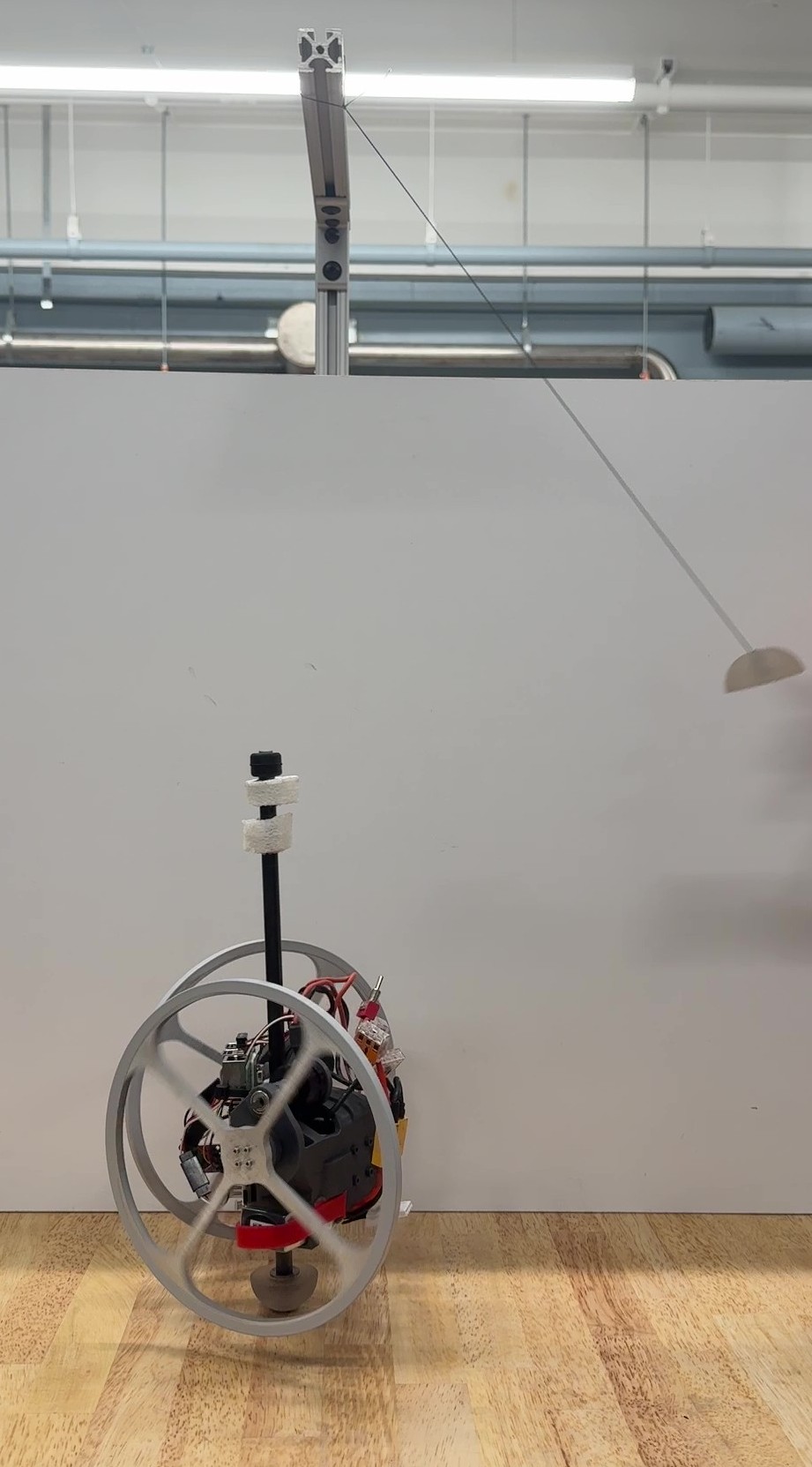}
    \caption{Test set-up for determining impulse rejection of the balance controller. Here the robot is hit by a mass at the end on a pendulum against the highest point of the robot.}
    \label{8}
\end{figure}
\begin{figure}
    \centering
    \includegraphics[width=1\linewidth]{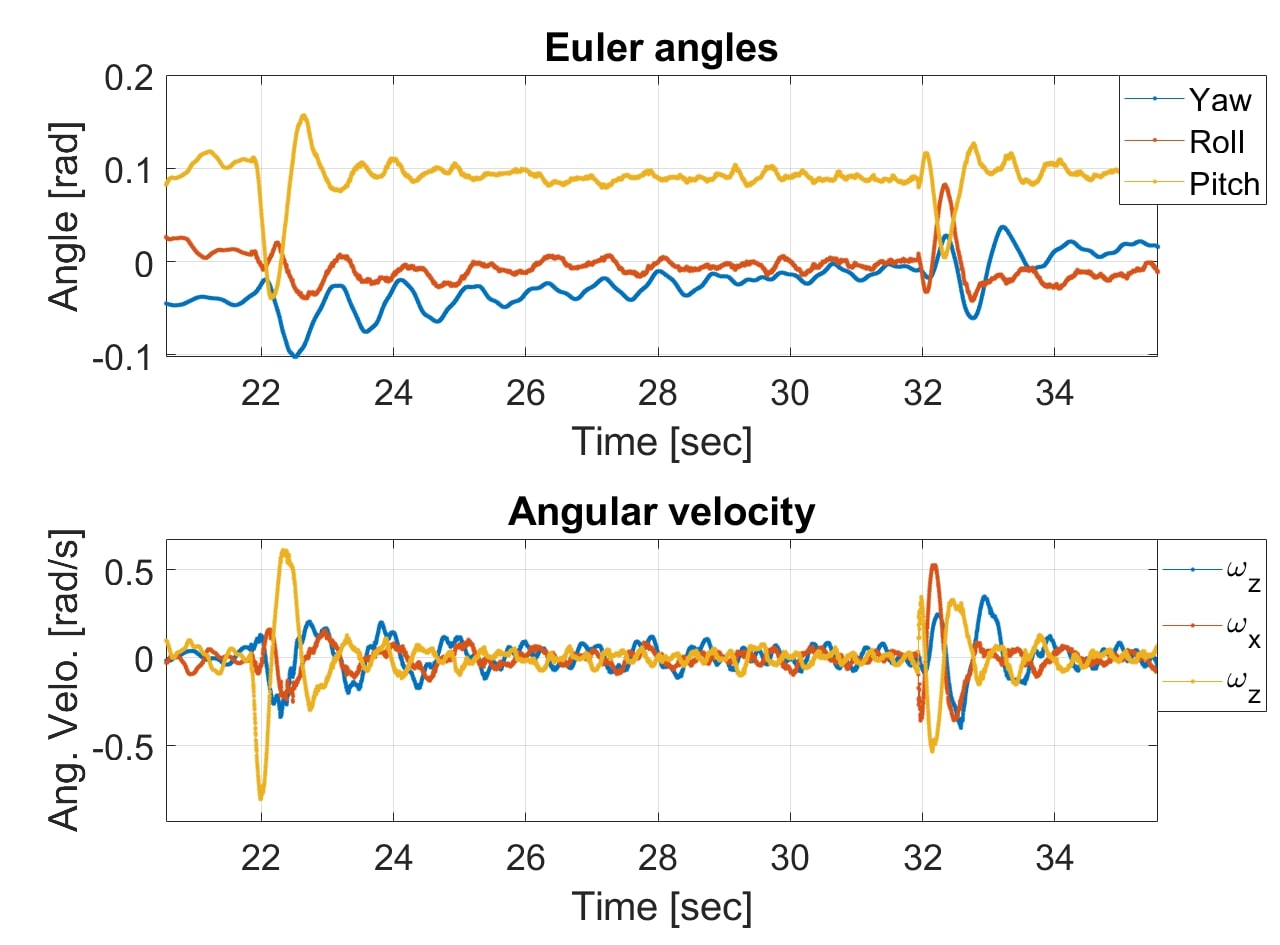}
    \caption{Euler angles and angular velocity of the robot in balance disturbance experiments. A pendulum struck the robot in the front-back direction at 22 and 32 s with impulse magnitudes of $0.013~N m s$ and $0.009~N m s$ respectively. The offsets from $0$ in the Euler angles are due to the balance offset of the robot since the center of mass is not located exactly along the leg.}
    \label{9}
\end{figure}
\subsection{Disturbance rejection while balancing}
While the robot is balancing, it may encounter outside forces from unknown sources. The ability to resist and reject disturbances demonstrates the robustness of the balancing controller. Fig. \ref{9} displays known impact magnitudes to the robot and the respective responses. In these tests a pendulum swings into the body of the robot while it balances as shown in Fig. \ref{8}. Due to the cant angle of the wheels, the balance controller has greater control authority over the forward and back directions than it does over left right directions. 

In these experiments the balance controller rejected impulses of $0.01~N m s$ in both the forward-back and side-to-side directions. Fig. \ref{9} displays the response to an example trial of an impact to the top of the robot in the forward-back directions. The theoretical optimal control for this reaction wheel inverted pendulum system derived following \cite{driessen_experimental_2019} could reject a disturbance of $0.054$~$N m s$ but this theoretical value is not achieved due to measurement noise and other nonidealities.
\begin{figure}
    \centering
    \includegraphics[width=1\linewidth]{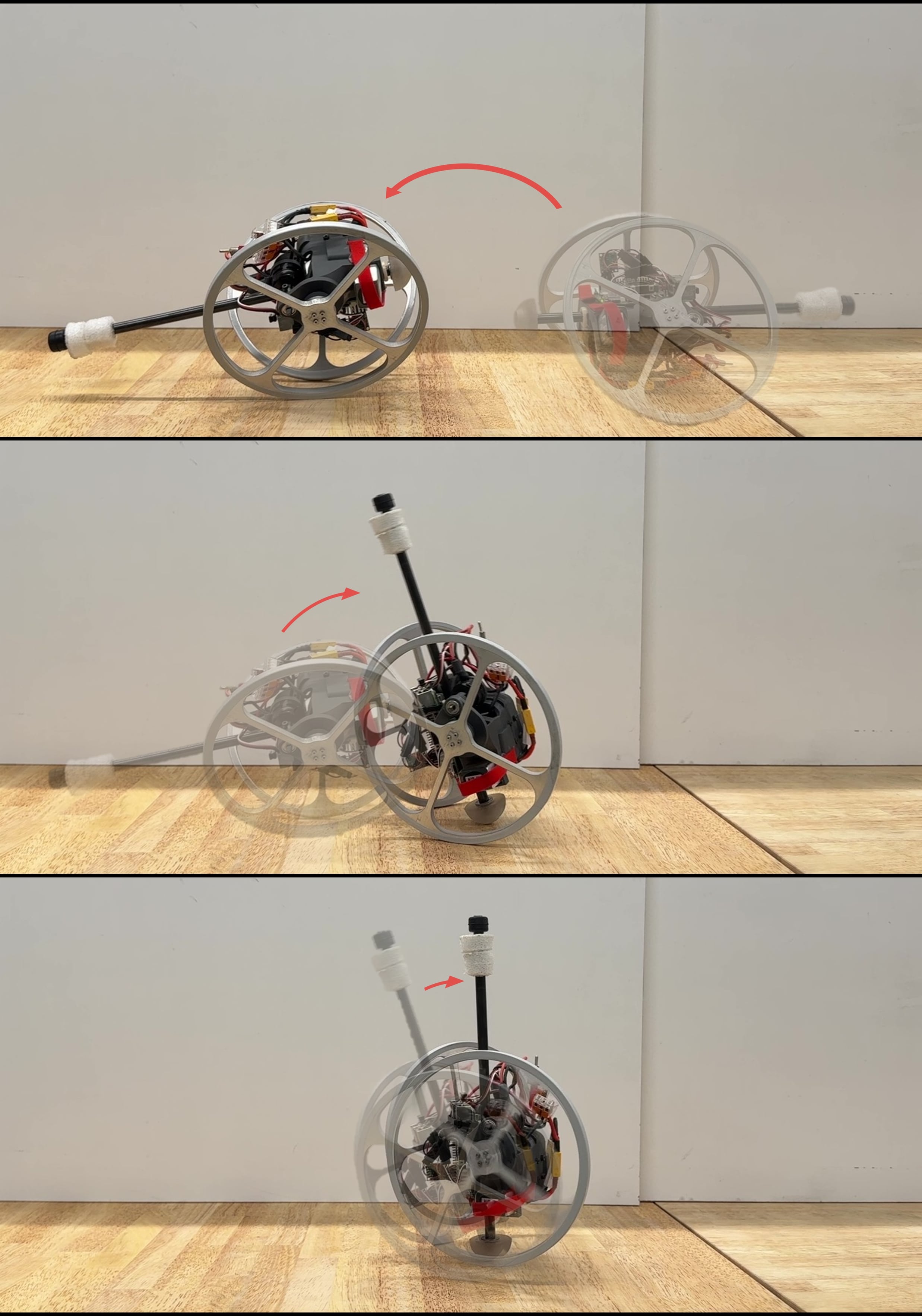}
    \caption{Overlay of self-righting from rolling to balancing. Top: Robot rolls forward and stops suddenly to flip over. Middle: Robot extends leg to shift weight backwards and tip upright. Bottom: Robot lifts itself off the ground and begins balancing. }
    \label{10}
\end{figure}
\begin{figure}
    \centering
    \includegraphics[width=1\linewidth]{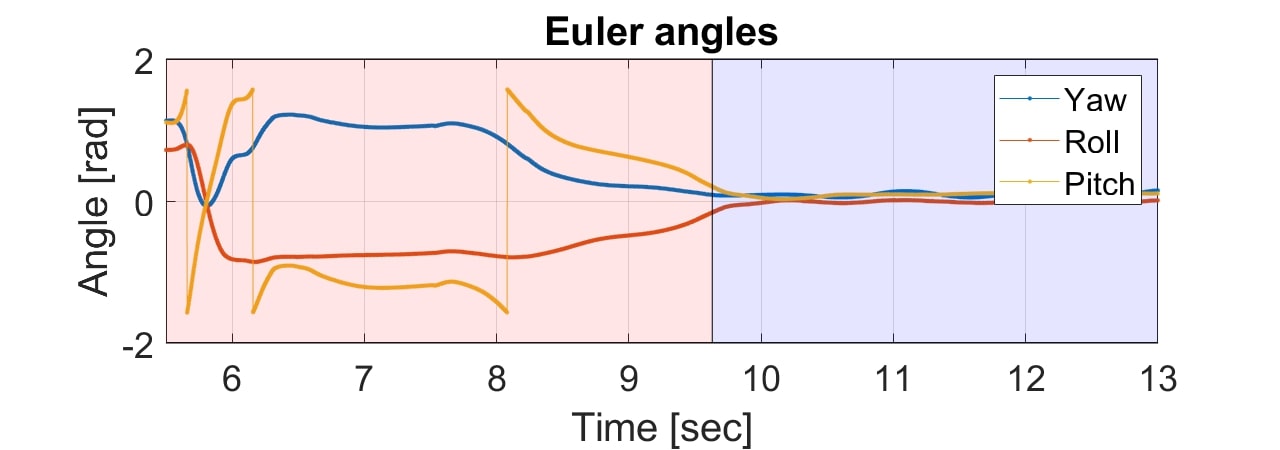}
    \caption{Euler angles during self-righting shown in figure \ref{10}. The red background denotes the system in rolling phase, and the blue background denotes the system in balancing phase. The discontinuity in pitch is due to the system passing the domain limits of $\pm\pi/2$.}
    \label{11}
\end{figure}
\subsection{Self-righting from rolling to balancing}
To transition from rolling to jumping, the robot must self-right, orienting its leg down and initiating balance control. This behavior also provides robustness to crashes after a jump, since if it ends in a fall, the robot can land on a side and roll from there or re-right itself and jump again.
Fig. \ref{10} depicts self-righting, and Fig. \ref{11} displays the associated Euler angles of the robot. To get itself upright, it brakes quickly while moving forward to flip over. Then it extends its leg to shift its mass so that it tips upward. Lastly, it stands up on its leg and begins balancing.

\begin{figure}
    \centering
    \includegraphics[width=1\linewidth]{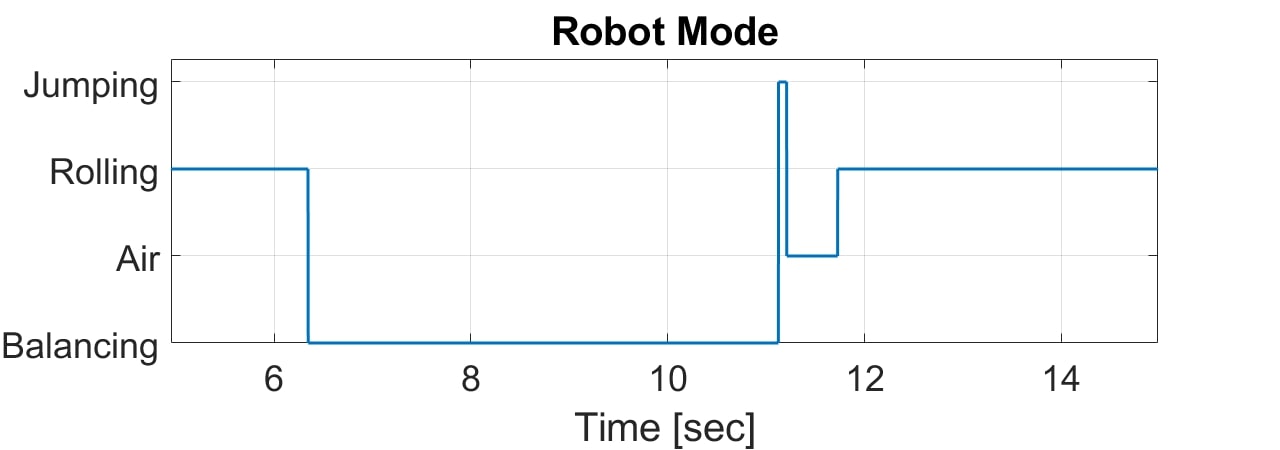}
    \caption{Robot Modes during the trial shown in figure \ref{12}. The robot goes from rolling to balancing, then to jumping into the air, and then landing and rolling again.}
    \label{13}
\end{figure}

\begin{figure}
    \centering
    \includegraphics[width=1\linewidth]{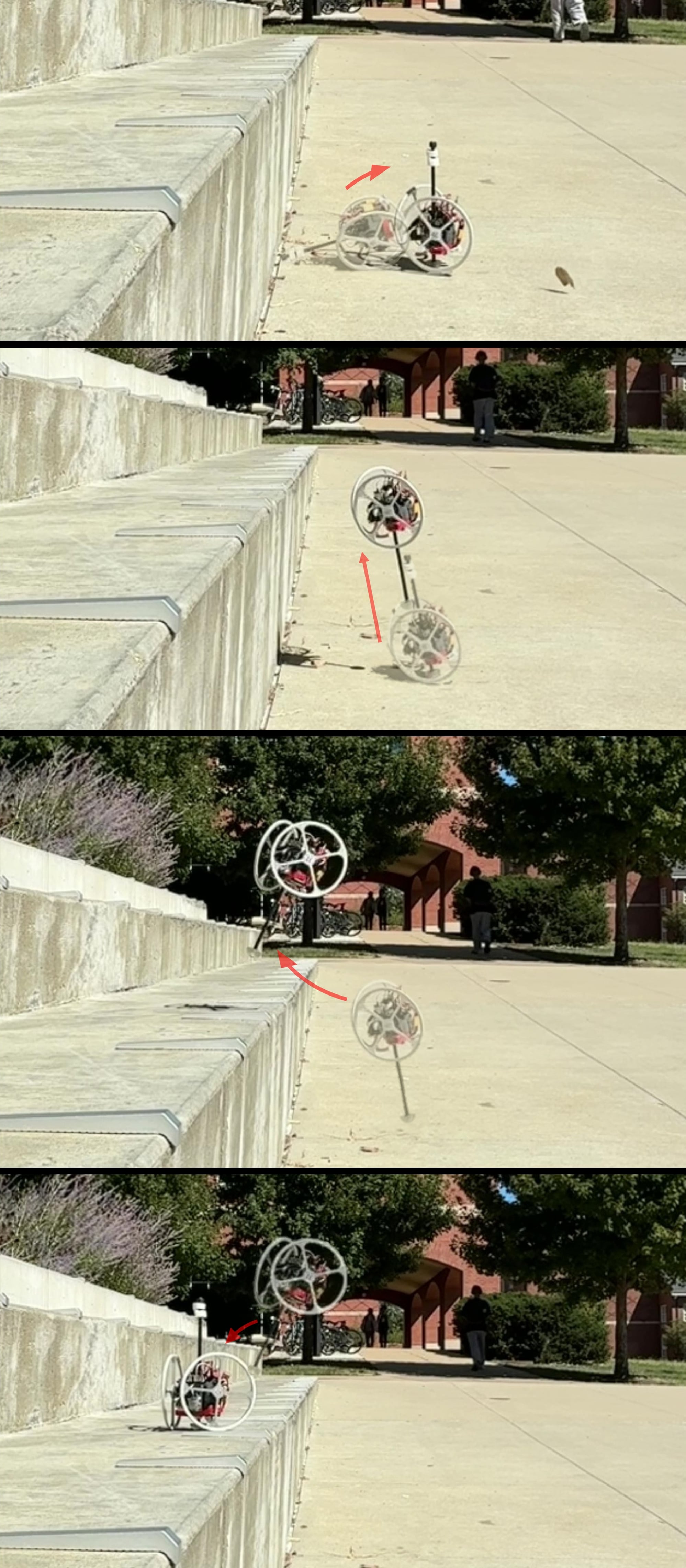}
    \caption{Overlay of combined behaviors: Rolling and self-righting into balancing mode, then leaning and jumping up. After jumping it reorients to point along the velocity vector, and finally lands softly, returning to rolling.}
    \label{12}
\end{figure}

The robot can combine the above behaviors using the methods described to navigate complicated environments. Fig. \ref{12} shows rolling and self-righting into the balance position at the base of a large step. Then, it leaps onto the step and damps out the impact of landing, perching itself on top of the ledge ready to roll away or initiate another leap. Fig. \ref{13} shows the state machine states throughout the demonstration sequence through rolling, balancing, jumping, flight, and rolling again.

\section{Conclusion and Future Work}
In this work we introduce a new robotic platform that despite underactuation can perform rolling and targeted jumping locomotion. This platform is developing ways that future space robots may explore the rough terrain and exotic conditions presented on extraterrestrial bodies, such as Enceladus. We also present two controllers that allow the system to balance on a point and reorient in the air while using only two reaction wheels in a 3D space. This allows the system to perform complex maneuvers while minimizing mechanical complexity. Specifically, we demonstrate the ability to roll, balance, self-right, and jump while having only three actuators in the entire system. Each of these abilities show the capability of this platform and its potential. However, the further refinement of the aerial controller and landing is required to make a balanced upright landing.

For the future of the controllers introduced in this paper, several expansions can be made. The self-righting control can be adapted for more robust performance on a variety of surfaces.
The aerial controller can also be expanded upon using different controller techniques other than a simple PD control, bringing the system to its desired state in finite time. 

In the future we plan to further develop the system for extraterrestrial exploration by optimizing the efficiency of the systems, replacing the leg with a linkage that better suits the exploration domain, and introducing new maneuvers such as wall jumps and instrument pointing to further increase this platform's ability. We also plan to test jumping on granular media to experimentally determine a foot profile design to jump effectively on the surfaces like Enceladus' ice particles. Additionally, the inclusion of vision capabilities will allow this platform to scout areas that may be desirable for exploration or safe to land and jump again. Lastly, a future system must be fitted with scientific instruments so that scientific duties can be performed. 

\section*{Acknowledgment}

The authors would like to thank their team in the NASA Innovative Advanced Concepts project LEAP - Legged Exploration Across the Plume for their collaboration: Ethan Schaler, Morgan Cable, Stephen Gerdts, Colin Creager, and Ryan McCormic.

\bibliographystyle{IEEEtran}
\bibliography{References}

@incollection{zelinka_attitude_2020,
	address = {Cham},
	title = {Attitude {Control} of {Jumping} {Robot} with {Bending}-{Stretching} {Mechanism}},
	volume = {554},
	isbn = {978-3-030-14906-2 978-3-030-14907-9},
	url = {http://link.springer.com/10.1007/978-3-030-14907-9_85},
	language = {en},
	urldate = {2023-12-11},
	booktitle = {{AETA} 2018 - {Recent} {Advances} in {Electrical} {Engineering} and {Related} {Sciences}: {Theory} and {Application}},
	publisher = {Springer International Publishing},
	author = {Ong, Chea Xin and Nomura, Yurika and Ishikawa, Jun},
	editor = {Zelinka, Ivan and Brandstetter, Pavel and Trong Dao, Tran and Hoang Duy, Vo and Kim, Sang Bong},
	year = {2020},
	doi = {10.1007/978-3-030-14907-9_85},
	note = {Series Title: Lecture Notes in Electrical Engineering},
	pages = {883--893},
}

@inproceedings{nomura_attitude_2018,
	address = {Enschede},
	title = {Attitude {Control} for {Underactuated} {Hopping} {Robots} {Using} {Nonlinear} {Output} {Zeroing} {Controller}},
	isbn = {978-1-5386-8183-1},
	url = {https://ieeexplore.ieee.org/document/8487770/},
	doi = {10.1109/BIOROB.2018.8487770},
	urldate = {2023-12-11},
	booktitle = {2018 7th {IEEE} {International} {Conference} on {Biomedical} {Robotics} and {Biomechatronics} ({Biorob})},
	publisher = {IEEE},
	author = {Nomura, Yurika and Ishikawa, Jun},
	month = aug,
	year = {2018},
	pages = {1120--1126},
}

@inproceedings{kovac_miniature_2008,
	address = {Pasadena, CA, USA},
	title = {A miniature 7g jumping robot},
	isbn = {978-1-4244-1646-2},
	url = {http://ieeexplore.ieee.org/document/4543236/},
	doi = {10.1109/ROBOT.2008.4543236},
	language = {en},
	urldate = {2023-12-11},
	booktitle = {2008 {IEEE} {International} {Conference} on {Robotics} and {Automation}},
	publisher = {IEEE},
	author = {Kovac, Mirko and Fuchs, Martin and Guignard, Andre and Zufferey, Jean-Christophe and Floreano, Dario},
	month = may,
	year = {2008},
	pages = {373--378},
}

@inproceedings{jianguo_zhao_controlling_2013,
	address = {Tokyo},
	title = {Controlling aerial maneuvering of a miniature jumping robot using its tail},
	isbn = {978-1-4673-6358-7 978-1-4673-6357-0},
	url = {http://ieeexplore.ieee.org/document/6696900/},
	doi = {10.1109/IROS.2013.6696900},
	language = {en},
	urldate = {2023-12-11},
	booktitle = {2013 {IEEE}/{RSJ} {International} {Conference} on {Intelligent} {Robots} and {Systems}},
	publisher = {IEEE},
	author = {{Jianguo Zhao} and {Tianyu Zhao} and {Ning Xi} and Cintron, Fernando J. and Mutka, Matt W. and {Li Xiao}},
	month = nov,
	year = {2013},
	pages = {3802--3807},
}

@inproceedings{zhao_miniature_2014,
	address = {Hong Kong, China},
	title = {A miniature 25 grams running and jumping robot},
	isbn = {978-1-4799-3685-4},
	url = {http://ieeexplore.ieee.org/document/6907609/},
	doi = {10.1109/ICRA.2014.6907609},
	language = {en},
	urldate = {2023-12-11},
	booktitle = {2014 {IEEE} {International} {Conference} on {Robotics} and {Automation} ({ICRA})},
	publisher = {IEEE},
	author = {Zhao, Jianguo and Yan, Weihan and Xi, Ning and Mutka, Matt W. and Xiao, Li},
	month = may,
	year = {2014},
	pages = {5115--5120},
}

@article{muehlebach_nonlinear_2017,
	title = {Nonlinear {Analysis} and {Control} of a {Reaction}-{Wheel}-{Based} 3-{D} {Inverted} {Pendulum}},
	volume = {25},
	issn = {1063-6536, 1558-0865},
	url = {http://ieeexplore.ieee.org/document/7457287/},
	doi = {10.1109/TCST.2016.2549266},
	language = {en},
	number = {1},
	urldate = {2023-12-11},
	journal = {IEEE Transactions on Control Systems Technology},
	author = {Muehlebach, Michael and D'Andrea, Raffaello},
	month = jan,
	year = {2017},
	pages = {235--246},
}

@article{libby_comparative_2016,
	title = {Comparative {Design}, {Scaling}, and {Control} of {Appendages} for {Inertial} {Reorientation}},
	volume = {32},
	issn = {1552-3098, 1941-0468},
	url = {http://ieeexplore.ieee.org/document/7562541/},
	doi = {10.1109/TRO.2016.2597316},
	language = {en},
	number = {6},
	urldate = {2023-12-11},
	journal = {IEEE Transactions on Robotics},
	author = {Libby, Thomas and Johnson, Aaron M. and Chang-Siu, Evan and Full, Robert J. and Koditschek, Daniel E.},
	month = dec,
	year = {2016},
	pages = {1380--1398},
}

@article{chu_combining_2023,
	title = {Combining {Tail} and {Reaction} {Wheel} for {Underactuated} {Spatial} {Reorientation} in {Robot} {Falling} {With} {Quadratic} {Programming}},
	volume = {8},
	issn = {2377-3766, 2377-3774},
	url = {https://ieeexplore.ieee.org/document/10272254/},
	doi = {10.1109/LRA.2023.3322079},
	language = {en},
	number = {11},
	urldate = {2023-12-11},
	journal = {IEEE Robotics and Automation Letters},
	author = {Chu, Xiangyu and Wang, Shengzhi and Ng, Raymond and Fan, Chun Yin and An, Jiajun and Au, K. W. Samuel},
	month = nov,
	year = {2023},
	pages = {7783--7790},
}

@article{rui_nonlinear_2000,
	title = {Nonlinear attitude and shape control of spacecraft with articulated appendages and reaction wheels},
	volume = {45},
	issn = {00189286},
	url = {http://ieeexplore.ieee.org/document/871754/},
	doi = {10.1109/9.871754},
	language = {en},
	number = {8},
	urldate = {2023-12-11},
	journal = {IEEE Transactions on Automatic Control},
	author = {Rui, C. and Kolmanovsky, I.V. and McClamroch, N.H.},
	month = aug,
	year = {2000},
	pages = {1455--1469},
}

@inproceedings{klemm_ascento_2019,
	address = {Montreal, QC, Canada},
	title = {Ascento: {A} {Two}-{Wheeled} {Jumping} {Robot}},
	copyright = {https://ieeexplore.ieee.org/Xplorehelp/downloads/license-information/IEEE.html},
	isbn = {978-1-5386-6027-0},
	shorttitle = {Ascento},
	url = {https://ieeexplore.ieee.org/document/8793792/},
	doi = {10.1109/ICRA.2019.8793792},
	language = {en},
	urldate = {2025-09-10},
	booktitle = {2019 {International} {Conference} on {Robotics} and {Automation} ({ICRA})},
	publisher = {IEEE},
	author = {Klemm, Victor and Morra, Alessandro and Salzmann, Ciro and Tschopp, Florian and Bodie, Karen and Gulich, Lionel and Kung, Nicola and Mannhart, Dominik and Pfister, Corentin and Vierneisel, Marcus and Weber, Florian and Deuber, Robin and Siegwart, Roland},
	month = may,
	year = {2019},
	pages = {7515--7521},
}

@inproceedings{spiridonov_spacehopper_2024,
	address = {Yokohama, Japan},
	title = {{SpaceHopper}: {A} {Small}-{Scale} {Legged} {Robot} for {Exploring} {Low}-{Gravity} {Celestial} {Bodies}},
	copyright = {https://doi.org/10.15223/policy-029},
	isbn = {979-8-3503-8457-4},
	shorttitle = {{SpaceHopper}},
	url = {https://ieeexplore.ieee.org/document/10610057/},
	doi = {10.1109/ICRA57147.2024.10610057},
	language = {en},
	urldate = {2025-09-10},
	booktitle = {2024 {IEEE} {International} {Conference} on {Robotics} and {Automation} ({ICRA})},
	publisher = {IEEE},
	author = {Spiridonov, Alexander and Buehler, Fabio and Berclaz, Moriz and Schelbert, Valerio and Geurts, Jorit and Krasnova, Elena and Steinke, Emma and Toma, Jonas and Wuethrich, Joschua and Polat, Recep and Zimmermann, Wim and Arm, Philip and Rudin, Nikita and Kolvenbach, Hendrik and Hutter, Marco},
	month = may,
	year = {2024},
	pages = {3464--3470},
}

@article{featherstone_simple_2017,
	title = {A simple model of balancing in the plane and a simple preview balance controller},
	volume = {36},
	issn = {0278-3649, 1741-3176},
	url = {https://journals.sagepub.com/doi/10.1177/0278364917691114},
	doi = {10.1177/0278364917691114},
	language = {en},
	number = {13-14},
	urldate = {2025-09-10},
	journal = {The International Journal of Robotics Research},
	author = {Featherstone, Roy},
	month = dec,
	year = {2017},
	pages = {1489--1507},
}

@article{featherstone_quantitative_2016,
	title = {Quantitative measures of a robot’s physical ability to balance},
	volume = {35},
	issn = {0278-3649, 1741-3176},
	url = {https://journals.sagepub.com/doi/10.1177/0278364916669599},
	doi = {10.1177/0278364916669599},
	language = {en},
	number = {14},
	urldate = {2025-09-10},
	journal = {The International Journal of Robotics Research},
	author = {Featherstone, Roy},
	month = dec,
	year = {2016},
	pages = {1681--1696},
}

@inproceedings{driessen_experimental_2019,
	address = {Montreal, QC, Canada},
	title = {Experimental {Demonstration} of {High}-{Performance} {Robotic} {Balancing}},
	copyright = {https://ieeexplore.ieee.org/Xplorehelp/downloads/license-information/IEEE.html},
	isbn = {978-1-5386-6027-0},
	url = {https://ieeexplore.ieee.org/document/8794447/},
	doi = {10.1109/ICRA.2019.8794447},
	language = {en},
	urldate = {2025-09-11},
	booktitle = {2019 {International} {Conference} on {Robotics} and {Automation} ({ICRA})},
	publisher = {IEEE},
	author = {Driessen, Josephus J. M. and Gkikakis, Antonios E. and Featherstone, Roy and Singh, B. Roodra P.},
	month = may,
	year = {2019},
	pages = {9459--9465},
}

@inproceedings{haldane_repetitive_2017,
	address = {Vancouver, BC},
	title = {Repetitive extreme-acceleration (14-g) spatial jumping with {Salto}-{1P}},
	isbn = {978-1-5386-2682-5},
	url = {http://ieeexplore.ieee.org/document/8206172/},
	doi = {10.1109/IROS.2017.8206172},
	language = {en},
	urldate = {2025-09-11},
	booktitle = {2017 {IEEE}/{RSJ} {International} {Conference} on {Intelligent} {Robots} and {Systems} ({IROS})},
	publisher = {IEEE},
	author = {Haldane, Duncan W. and Yim, Justin K. and Fearing, Ronald S.},
	month = sep,
	year = {2017},
	pages = {3345--3351},
}

@article{burns_design_2025,
	title = {Design and {Control} of a {High}-{Performance} {Hopping} {Robot}},
	volume = {10},
	copyright = {https://ieeexplore.ieee.org/Xplorehelp/downloads/license-information/IEEE.html},
	issn = {2377-3766, 2377-3774},
	url = {https://ieeexplore.ieee.org/document/10964875/},
	doi = {10.1109/LRA.2025.3560884},
	language = {en},
	number = {6},
	urldate = {2025-09-11},
	journal = {IEEE Robotics and Automation Letters},
	author = {Burns, Samuel and Woodward, Matthew},
	month = jun,
	year = {2025},
	pages = {5641--5648},
}

@inproceedings{gajamohan_cubli_2012,
	address = {Vilamoura-Algarve, Portugal},
	title = {The {Cubli}: {A} cube that can jump up and balance},
	isbn = {978-1-4673-1736-8 978-1-4673-1737-5 978-1-4673-1735-1},
	shorttitle = {The {Cubli}},
	url = {http://ieeexplore.ieee.org/document/6385896/},
	doi = {10.1109/IROS.2012.6385896},
	language = {en},
	urldate = {2025-09-11},
	booktitle = {2012 {IEEE}/{RSJ} {International} {Conference} on {Intelligent} {Robots} and {Systems}},
	publisher = {IEEE},
	author = {Gajamohan, Mohanarajah and Merz, Michael and Thommen, Igor and D'Andrea, Raffaello},
	month = oct,
	year = {2012},
	pages = {3722--3727},
}

@inproceedings{kalita_dynamics_2020,
	address = {Big Sky, MT, USA},
	title = {Dynamics and {Control} of a {Hopping} {Robot} for {Extreme} {Environment} {Exploration} on the {Moon} and {Mars}},
	copyright = {https://ieeexplore.ieee.org/Xplorehelp/downloads/license-information/IEEE.html},
	isbn = {978-1-7281-2734-7},
	url = {https://ieeexplore.ieee.org/document/9172617/},
	doi = {10.1109/AERO47225.2020.9172617},
	language = {en},
	urldate = {2025-09-11},
	booktitle = {2020 {IEEE} {Aerospace} {Conference}},
	publisher = {IEEE},
	author = {Kalita, Himangshu and Gholap, Akash S. and Thangavelautham, Jekan},
	month = mar,
	year = {2020},
	pages = {1--12},
}

@incollection{kulic_experimental_2017,
	address = {Cham},
	title = {Experimental {Methods} for {Mobility} and {Surface} {Operations} of {Microgravity} {Robots}},
	volume = {1},
	copyright = {http://www.springer.com/tdm},
	isbn = {978-3-319-50114-7 978-3-319-50115-4},
	url = {http://link.springer.com/10.1007/978-3-319-50115-4_65},
	language = {en},
	urldate = {2025-09-11},
	booktitle = {2016 {International} {Symposium} on {Experimental} {Robotics}},
	publisher = {Springer International Publishing},
	author = {Hockman, Benjamin and Reid, Robert G. and Nesnas, Issa A. D. and Pavone, Marco},
	editor = {Kulić, Dana and Nakamura, Yoshihiko and Khatib, Oussama and Venture, Gentiane},
	year = {2017},
	doi = {10.1007/978-3-319-50115-4_65},
	note = {Series Title: Springer Proceedings in Advanced Robotics},
	pages = {752--763},
}

@article{raibert_hopping_1984,
	title = {Hopping in legged systems — {Modeling} and simulation for the two-dimensional one-legged case},
	volume = {SMC-14},
	issn = {0018-9472, 2168-2909},
	url = {http://ieeexplore.ieee.org/document/6313238/},
	doi = {10.1109/TSMC.1984.6313238},
	language = {en},
	number = {3},
	urldate = {2025-09-11},
	journal = {IEEE Transactions on Systems, Man, and Cybernetics},
	author = {Raibert, Marc H.},
	month = may,
	year = {1984},
	pages = {451--463},
}

@article{zeglin_first_nodate,
	title = {First {Hops} of the {3D} {Bow} {Leg}},
	language = {en},
	author = {Zeglin, Garth and Brown, H Benjamin},
        year = {2002},
        journal={Proceedings of International Conference on Climbing and Walking Robots},
        pages={357--364},
}

@article{choukroun_sampling_2021,
	title = {Sampling {Plume} {Deposits} on {Enceladus}’ {Surface} to {Explore} {Ocean} {Materials} and {Search} for {Traces} of {Life} or {Biosignatures}},
	volume = {2},
	issn = {2632-3338},
	url = {https://iopscience.iop.org/article/10.3847/PSJ/abf2c5},
	doi = {10.3847/PSJ/abf2c5},
	language = {en},
	number = {3},
	urldate = {2025-09-11},
	journal = {The Planetary Science Journal},
	author = {Choukroun, Mathieu and Backes, Paul and Cable, Morgan L. and Fayolle, Edith C. and Hodyss, Robert and Murdza, Andrii and Schulson, Erland M. and Badescu, Mircea and Malaska, Michael J. and Marteau, Eloïse and Molaro, Jamie L. and Moreland, Scott J. and Noell, Aaron C. and Nordheim, Tom A. and Okamoto, Tyler and Riccobono, Dario and Zacny, Kris},
	month = jun,
	year = {2021},
	pages = {100},
}

@article{mackenzie_enceladus_2021,
	title = {The {Enceladus} {Orbilander} {Mission} {Concept}: {Balancing} {Return} and {Resources} in the {Search} for {Life}},
	volume = {2},
	issn = {2632-3338},
	shorttitle = {The {Enceladus} {Orbilander} {Mission} {Concept}},
	url = {https://iopscience.iop.org/article/10.3847/PSJ/abe4da},
	doi = {10.3847/PSJ/abe4da},
	language = {en},
	number = {2},
	urldate = {2025-09-11},
	journal = {The Planetary Science Journal},
	author = {MacKenzie, Shannon M. and Neveu, Marc and Davila, Alfonso F. and Lunine, Jonathan I. and Craft, Kathleen L. and Cable, Morgan L. and Phillips-Lander, Charity M. and Hofgartner, Jason D. and Eigenbrode, Jennifer L. and Waite, J. Hunter and Glein, Christopher R. and Gold, Robert and Greenauer, Peter J. and Kirby, Karen and Bradburne, Christopher and Kounaves, Samuel P. and Malaska, Michael J. and Postberg, Frank and Patterson, G. Wesley and Porco, Carolyn and Núñez, Jorge I. and German, Chris and Huber, Julie A. and McKay, Christopher P. and De Vera, Jean-Pierre and Brucato, John Robert and Spilker, Linda J.},
	month = apr,
	year = {2021},
	pages = {77},
}

@article{des_marais_nasa_2008,
	title = {The {NASA} {Astrobiology} {Roadmap}},
	volume = {8},
	issn = {1531-1074, 1557-8070},
	url = {http://www.liebertpub.com/doi/10.1089/ast.2008.0819},
	doi = {10.1089/ast.2008.0819},
	language = {en},
	number = {4},
	urldate = {2025-09-11},
	journal = {Astrobiology},
	author = {Des Marais, David J. and Nuth, Joseph A. and Allamandola, Louis J. and Boss, Alan P. and Farmer, Jack D. and Hoehler, Tori M. and Jakosky, Bruce M. and Meadows, Victoria S. and Pohorille, Andrew and Runnegar, Bruce and Spormann, Alfred M.},
	month = aug,
	year = {2008},
	pages = {715--730},
}

@article{cable_science_2021,
	title = {The {Science} {Case} for a {Return} to {Enceladus}},
	volume = {2},
	issn = {2632-3338},
	url = {https://iopscience.iop.org/article/10.3847/PSJ/abfb7a},
	doi = {10.3847/PSJ/abfb7a},
	language = {en},
	number = {4},
	urldate = {2025-09-11},
	journal = {The Planetary Science Journal},
	author = {Cable, Morgan L. and Porco, Carolyn and Glein, Christopher R. and German, Christopher R. and MacKenzie, Shannon M. and Neveu, Marc and Hoehler, Tori M. and Hofmann, Amy E. and Hendrix, Amanda R. and Eigenbrode, Jennifer and Postberg, Frank and Spilker, Linda J. and McEwen, Alfred and Khawaja, Nozair and Hunter Waite, J. and Wurz, Peter and Helbert, Jörn and Anbar, Ariel and De Vera, Jean-Pierre and Núñez, Jorge},
	month = aug,
	year = {2021},
	pages = {132},
}

@article{ackerman2012boston,
  title={Boston dynamics sand flea robot demonstrates astonishing jumping skills},
  author={Ackerman, Evan},
  journal={IEEE Spectrum Robotics Blog},
  volume={2},
  number={1},
  pages={1},
  year={2012}
}

@misc{david2005opportunity,
  title={Opportunity Mars Rover stuck in sand},
  author={David, Leonard},
  year={2005}
}

@article{vaquero2024eels,
  title={EELS: Autonomous snake-like robot with task and motion planning capabilities for ice world exploration},
  author={Vaquero, Tiago Stegun and Daddi, Guglielmo and Thakker, Rohan and Paton, Michael and Jasour, A and Strub, Marlin P and Swan, R Michael and Royce, Rob and Gildner, Matthew and Tosi, P and others},
  journal={Science robotics},
  volume={9},
  number={88},
  pages={eadh8332},
  year={2024},
  publisher={American Association for the Advancement of Science}
}
\addtolength{\textheight}{-12cm}   
\newpage
\end{document}